\documentclass{clv3}

\usepackage{url}
\usepackage{latexsym}

\usepackage[utf8]{inputenc} 
\usepackage[T1]{fontenc}    
\usepackage{hyperref}       
\usepackage{url}            
\usepackage{booktabs}       
\usepackage{amsfonts}       
\usepackage{nicefrac}       
\usepackage{microtype}      

\usepackage{graphicx}
\usepackage{algorithm}
\usepackage[noend]{algpseudocode}
\usepackage{subcaption}

\usepackage{amssymb, amsmath}
\usepackage{fixltx2e}

\usepackage{calc}
\usepackage[table,xcdraw]{xcolor}

\def\R{\mathbb{R}}

\def\cm#1{{\textsc{#1}_\circ}}
\def\cmt#1{{\textsc{#1}_\circ^\transpose}}

\def\invsc#1{[#1]^{\ominus}}
\def\sc#1{[#1]^{\oplus}}

\def\cconv{\ast}

\def\shufcconv{\circledast}

\def\ffbox#1{\resizebox{20pt}{20pt}{\vbox to 15pt {\vfil
\hbox to 2.5em{\fbox{{#1}}}%
\vfil
}}}

\def\tensor#1{\textbf{#1}}

\def\matrix#1{\mathbf{#1}}

\def\cykcell#1#2{\setlength{\tabcolsep}{.16667em}\begin{tabular}{|p{0.5cm}|}\cellcolor{gray!30}#2\\\cellcolor{gray!50}{\tiny #1}\end{tabular}}

\def\symbA{\hspace{-2mm}\begin{tabular}{c}{\tiny a}\\\ominoc[magenta]{..*\\.**\\***\\***\\..*}\end{tabular}\hspace{-2mm}}
\def\rsymbA{\hspace{-2mm}\begin{tabular}{c}{\tiny a}\\\ominoc[magenta]{***\\**\\*\\*\\***}\end{tabular}\hspace{-2mm}}
\def\symbB{\hspace{-2mm}\begin{tabular}{c}{\tiny b}\\\ominoc[cyan]{..*\\.**\\..*\\***\\..*}\end{tabular}\hspace{-2mm}}
\def\rsymbB{\hspace{-2mm}\begin{tabular}{c}{\tiny b}\\\ominoc[cyan]{***\\**\\***\\*\\***}\end{tabular}\hspace{-2mm}}
\def\symbD{\hspace{-2mm}\begin{tabular}{c}{\tiny D}\\\ominoc[olive]{.**\\..*\\.**\\..*\\.**}\end{tabular}\hspace{-2mm}}
\def\rsymbD{\hspace{-2mm}\begin{tabular}{c}{\tiny D}\\\rominoc[olive]{.**\\..*\\.**\\..*\\.**}\end{tabular}\hspace{-2mm}}
\def\symbE{\hspace{-2mm}\begin{tabular}{c}{\tiny E}\\\ominoc[green]{..*\\..*\\.**\\..*\\***}\end{tabular}\hspace{-2mm}}
\def\rsymbE{\hspace{-2mm}\begin{tabular}{c}{\tiny E}\\\rominoc[green]{..*\\..*\\.**\\..*\\***}\end{tabular}\hspace{-2mm}}
\def\symbS{\hspace{-2mm}\begin{tabular}{c}{\tiny S}\\\ominoc[yellow]{***\\..*\\.**\\..*\\***}\end{tabular}\hspace{-2mm}}
\def\rsymbS{\hspace{-2mm}\begin{tabular}{c}{\tiny S}\\\rominoc[yellow]{***\\..*\\.**\\..*\\***}\end{tabular}\hspace{-2mm}}
\def\indexZero{\hspace{-2mm}\begin{tabular}{c}{\tiny 0}\\\ominoc[violet]{***\\.**\\..*\\.**\\***}\end{tabular}\hspace{-2mm}}
\def\rindexZero{\hspace{-2mm}\begin{tabular}{c}{\tiny 0}\\\rominoc[violet]{***\\.**\\..*\\.**\\***}\end{tabular}\hspace{-2mm}}
\def\indexOne{\hspace{-2mm}\begin{tabular}{c}{\tiny 1}\\\ominoc[red]{***\\..*\\..*\\..*\\***}\end{tabular}\hspace{-2mm}}
\def\rindexOne{\hspace{-2mm}\begin{tabular}{c}{\tiny 1}\\\rominoc[red]{***\\..*\\..*\\..*\\***}\end{tabular}\hspace{-2mm}}
\def\indexTwo{\hspace{-2mm}\begin{tabular}{c}{\tiny 2}\\\ominoc[teal]{***\\.**\\.**\\..*\\***}\end{tabular}\hspace{-2mm}}
\def\indexThree{\hspace{-2mm}\begin{tabular}{c}{\tiny 3}\\\ominoc[blue]{***\\..*\\.**\\..*\\..*}\end{tabular}\hspace{-2mm}}
\def\rindexTwo{\hspace{-2mm}\begin{tabular}{c}{\tiny 2}\\\rominoc[teal]{***\\.**\\.**\\..*\\***}\end{tabular}\hspace{-2mm}}
\def\rindexThree{\hspace{-2mm}\begin{tabular}{c}{\tiny 3}\\\rominoc[blue]{***\\..*\\.**\\..*\\..*}\end{tabular}\hspace{-2mm}}

\makeatletter
\def\omino#1{{%
\setlength{\fboxsep}{0pt}
\unitlength2\p@
\@tempcnta\z@
\@tempcntb\@ne
\count@\z@
\xomino#1\relax
\ffbox{\begin{picture}(\@tempcnta,\@tempcntb)(0,-\@tempcntb)%
\@tempcnta\z@
\@tempcntb\@ne
\count@\z@
\xxomino#1\relax
\end{picture}}%
}%
}

\def\xomino#1{%
\ifx\relax#1%
\else
\ifx\\#1%
\ifnum\count@>\@tempcnta \@tempcnta\count@\fi
\advance\@tempcntb\@ne
\count@\z@
\else
\advance\count@\@ne
\fi
\expandafter\xomino
\fi}

\def\xxomino#1{%
\ifx\relax#1%
\else
  \ifx\\#1%
    \advance\@tempcntb\@ne
    \count@\z@
  \else
    \advance\count@\@ne
    \ifx*#1%
      \put(\count@,-\@tempcntb){\kern-2pt\rule{2pt}{2pt}}%
    \fi
  \fi
\expandafter\xxomino
\fi}

\makeatletter
\def\romino#1{{%
\setlength{\fboxsep}{0pt}
\unitlength2\p@
\@tempcnta\z@
\@tempcntb\@ne
\count@\z@
\xromino#1\relax
\ffbox{\begin{picture}(\@tempcnta,\@tempcntb)(0,-\@tempcntb)%
\@tempcnta\z@
\@tempcntb\@ne
\count@\z@
\xxromino#1\relax
\end{picture}}%
}%
}

\def\xromino#1{%
\ifx\relax#1%
\else
\ifx\\#1%
\ifnum\count@>\@tempcnta \@tempcnta\count@\fi
\advance\@tempcntb\@ne
\count@\z@
\else
\advance\count@\@ne
\fi
\expandafter\xromino
\fi}

\def\xxromino#1{%
\ifx\relax#1%
\else
  \ifx\\#1%
    \advance\@tempcntb\@ne
    \count@\z@
  \else
    \advance\count@\@ne
    \ifx.#1%
      \put(\count@,-\@tempcntb){\kern-2pt\rule{2pt}{2pt}}%
    \fi
  \fi
\expandafter\xxromino
\fi}
\makeatother

\newcommand{\eat}[1]{}
\newcommand{\gcomment}[1]{{\color{red} {\bf {#1}}}}
\newcommand{\termdef}[1]{\textbf{{#1}}}
\newcommand{\mleft}{\mathit{left}}
\newcommand{\mright}{\mathit{right}}
\newcommand{\rmatrix}[2]{\matrix{R}_{{#1}}^{({#2})}}
\newcommand{\transpose}{{\rm{T}}}

\begin{document}

\title{CYK Parsing over Distributed Representations}

\author{Fabio Massimo Zanzotto\thanks{%
  University of Rome Tor Vergata, 
  Viale del Politecnico 1, 
  00133 Roma. E-mail:~\texttt{fabio.massimo.zanzotto@uniroma2.it}}}
\affil{University of Rome Tor Vergata}

\author{Giordano Cristini\thanks{%
  University of Rome Tor Vergata, 
  Viale del Politecnico 1, 
  00133 Roma. E-mail:~\texttt{giordano.cristini@gmail.com}}}
\affil{University of Rome Tor Vergata}

\author{Giorgio Satta\thanks{%
    Department of Information Engineering, University of Padua, via
    Gradenigo 6/A, 35131 Padova,
    Italy. E-mail:~\texttt{satta@dei.unipd.it}}}
\affil{University of Padua}

\runningauthor{Zanzotto, Cristini, Satta}
\runningtitle{Distributed CYK Parsing}

\maketitle

\begin{abstract}

Syntactic parsing is a key task in natural language processing.  This task has been dominated by symbolic, grammar-based parsers. Neural networks, with their distributed representations, are challenging these methods. 
In this article we show that existing symbolic parsing algorithms can cross the border and be entirely formulated over distributed representations.  To this end we introduce a version of the traditional Cocke-Younger-Kasami (CYK) algorithm, called D-CYK, which is entirely defined over distributed representations. Our D-CYK uses matrix multiplication on real number matrices of size independent of the length of the input string. These operations are compatible with traditional neural networks. Experiments show that our D-CYK approximates the original CYK algorithm.
By showing that CYK can be entirely performed on distributed representations, we open the way to the definition of recurrent layers of CYK-informed neural networks. 

\end{abstract}

\section{Introduction}
\label{intro}

The area of natural language parsing has been dominated for decades by the so-called symbolic paradigm of artificial intelligence, which embraces the collection of methods that are based on high-level, human-readable, symbolic representations.  The Cocke-Younger-Kasami algorithm (CYK) \cite{Cocke:1969:PLC:1097042,Younger1967,Kasami:1965:CYK}, the Early algorithm \cite{Earley:1970:ECP:362007.362035} and the shift-reduce  algorithm \cite{Sippu:88} are at the core of most common algorithms for natural language parsing, both constituency-based and dependency-based, and they all use symbolic representations for grammar rules, parser states and syntactic trees. 

Starting with the early 90's and the surge of data-driven methods in natural language processing, grammar rules and their weights have been estimated from large data sets of syntactically annotated sentences, and probabilistic parsers and parsers based on discriminative models have flourished.  Nonetheless, all these methods are based on the above mentioned parsing algorithms, with grammar rules and parser states still symbolically represented.  With the recent development of deep learning techniques, fast and accurate dependency  parsers have been designed on top of the shift-reduce algorithm, where parser actions are parameterized using neural networks~\cite{ChenM14,AmbatiDS16}. Furthermore, the use of recurrent neural networks has made it possible to design models that condition parsing actions on the entire syntactic derivation history~\cite{Titov:2007,Dyer2016}, resulting in considerable enhancement of parsing performance.  Still, for all of these parsers based on neural networks, the underlying configuration of the parser is a stack of symbols and the search space is represented at that level.

The main goal in the above mentioned research is to inform parsing decisions using contextual information as rich as possible.  This means that the context-free assumption of the underlying formalism is dropped.  In case of the neural network parsers, this is achieved using distributed representations and matrix multiplication operations embedded into the underlying network architecture.  In this article we explore a somewhat tangential direction: we attempt to entirely remove the symbolic level of representation from the underlying parser.  There are at least two existing lines of research in this direction.  In a first line~\cite{Vinyals2014} parsing is seen as a sequence-to-sequence translation task, with the input sentence mapped to a linear syntactic interpretation, and training is done with long-short term memories (LSTM) networks over millions of sentences which have been annotated by existing parsers.  In this case it seems that the LSTM network learns two things: the associations among fragments of sentences and fragments of trees and a way of recombining these fragments in the final interpretation. The grammar representation is then hidden in the weights of the LSTM. 
In the second line of research~\cite{zanzotto-dellarciprete:2013:CVSC,Senay:2015:PES:2996831.2996847} both sentences and trees are represented in distributed vectors and neutral networks learn a way to map sentence vectors to tree vectors. However, the overall model does not have the ability to replicate distributed vectors for trees from distributed vectors for sentences. Resulting vectors are not accurate enough to have an impact on final tasks~\cite{zanzotto-dellarciprete:2013:CVSC}. Moreover, in this case as well the grammar is represented in an unpredictable way in the weights of the multi-layer perceptron or the LSTM.

In this article we take a rather different route, and show that traditional parsing algorithms can cross the border of distributed representations. We propose a version of the CYK algorithm, called D-CYK, that entirely works on distributed representations. 
This is achieved by transforming the parsing table at the base of the CYK algorithm  into a real number square matrix, and by implementing the basic operations of the algorithm using matrix multiplication. Implementations of the CYK algorithm using matrix multiplications are well known in the literature~\cite{Graham:76,Valiant:75} but they use symbolic representations, while in our proposal grammar symbols as well as constituent indices are all encoded into real numbers and in a distributed way.  A second, important difference is that in the standard CYK algorithm the parsing table has size $(n+1) \times (n+1)$, with $n$ the length of the input string, while in our D-CYK the parsing table has size $d \times d$, where $d$ depends on the distributed representation.  This means that, up to some extent, we can parse input sentences of different lengths without changing the size of the parsing table in the D-CYK algorithm.  We are not aware of any parsing algorithm for context-free grammars having such property. 

We report experiments on toy grammars showing that our D-CYK can successfully approximate the CYK algorithm.  This novel approach opens the way to the investigation of parsing algorithms where the symbolic level of representation is entirely dropped.

\section{Preliminaries}

In this section we introduce the basics about the CYK algorithm and overview a class of distributed representation called holographic reduced representation. 

\begin{figure}
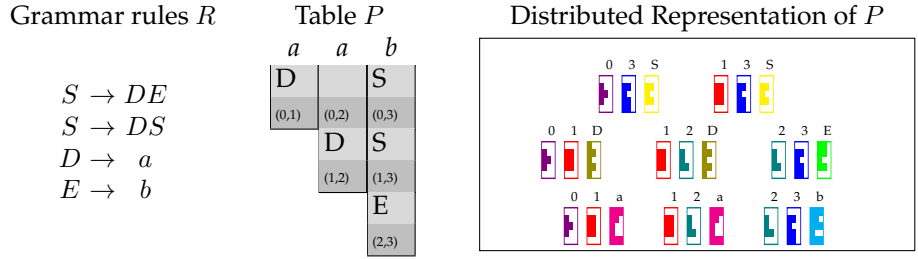

\begin{center}
\begin{tabular}{ccc}
Grammar rules $R$ & Table $P$ & Distributed Representation of $P$ \\
\begin{tabular}{c}
$\begin{array}{ccc}
S &\rightarrow & D E\\
S &\rightarrow & D S\\
D &\rightarrow & a\\
E &\rightarrow & b\\
\end{array}$
\end{tabular}
&
\begin{tabular}{c}
\setlength{\tabcolsep}{0.00001em}
\begin{tabular}{ccc}
\textit{a}    & \textit{a}    & \textit{b}    \\ 
\cline{1-3} 
\cykcell{(0,1)}{D} & \cykcell{(0,2)}{}   & \cykcell{(0,3)}{S}  \\ 
\cline{1-3} 
  & \cykcell{(1,2)}{D}   & \cykcell{(1,3)}{S}  \\ 
\cline{2-3} 
  &  & \cykcell{(2,3)}{E}  \\ 
\cline{3-3} 
\end{tabular}

\end{tabular}
&
\begin{tabular}{c}
\fbox{\parbox{\dimexpr\linewidth-75\fboxsep-2\fboxrule\relax}{\centering   
\begin{tabular}{ccc}
\rindexZero \rindexThree  \rsymbS
&
\rindexOne \rindexThree  \rsymbS
\end{tabular}\\
\begin{tabular}{ccc}
\rindexZero\rindexOne\rsymbD
&
\rindexOne\rindexTwo\rsymbD
&
\rindexTwo\rindexThree\rsymbE
\end{tabular}
\\
\begin{tabular}{cccc}
\rindexZero{}\rindexOne{}\rsymbA{}
&
\rindexOne{}\rindexTwo{}\rsymbA{}
&
\rindexTwo{}\rindexThree{}\rsymbB{}
\end{tabular}
}}
\end{tabular}
\end{tabular}
\caption{A simple context-free grammar, the CYK parsing table $P$ 
for the string $aab$, and the ``distributed'' representation 
$\matrix{P_{\mleft}}$ of $P$ in Tetris-like notation.}
\label{fig:cyk_matrix}
\end{center}
\end{figure}

\subsection{CYK algorithm}
\label{ssec:cyk}

The CYK algorithm is a classical algorithm for parsing based on context-free grammars, using dynamic programming.  We provide here a brief description of the algorithm in order to introduce the notation used in later sections; we closely follow the presentation in~\cite{Graham:76}.  The algorithm requires context-free grammars in Chomsky Normal Form (CNF), where each rule has the form $A \rightarrow B C$ or $A \rightarrow a$, where $A$, $B$, $C$ are nonterminal symbols and $a$ is a terminal symbol.  We write $R$ to denote the set of all rules of the grammar. 

Given an input string $w = a_1 \cdots a_n$, $n \geq 1$, where each $a_i$ is an alphabet symbol, the algorithm uses a 2-dimensional table $P$ of size $(n+1) \times (n+1)$, where each entry  stores a set of nonterminals representing partial parses of the input string.  More precisely, for $0 \leq i < j \leq n$,  a nonterminal $A$ belongs to set $P[i,j]$ if and only if there exists a parse tree with root $A$ generating the substring $a_{i+1} \cdots a_j$ of $w$.  Thus, $w$ can be parsed if the initial nonterminal of the grammar $S$ is added to $P[0,n]$.  
Algorithm~\ref{alg:CYK} shows how table $P$ is populated.  $P$ is first initialized using unary rules, at line~\ref{l:init}.  
Then each entry $P[i,j]$ is filled at line~9  
by looking at pairs $P[i,k]$ and $P[k,j]$ and by using binary rules. 

\begin{algorithm}
\begin{algorithmic}[1]
\For {$i \gets 1$ to $n$}
    \For {each $A \rightarrow a_i$ in $R$}
    	\State add $A$ to $P[i-1,i]$ \label{l:init}
    \EndFor
\EndFor
\For {$j \gets 2$ to $n$}
	\For {$i \gets j-2 $ to $0$}
		\For {$k \gets i+1$ to $j-1$}
    		\For {each $A \rightarrow B C$ in $R$}
				\If {$B \in P[i,k]$ and $C \in P[k,j]$} 
					\State add $A$ to $P[i,j]$ \label{l:binary}
				\EndIf 
    		\EndFor
    	\EndFor
    \EndFor
\EndFor
\end{algorithmic}
\caption{CYK(string $w = a_1 \cdots a_n$, rule set $R$) \Return table $P$}
\label{alg:CYK}
\end{algorithm}

A running example is presented in Figure~\ref{fig:cyk_matrix}, showing a set $R$ of grammar rules along with the table $P$ produced by the algorithm when processing the input string $w = aab$.  For instance, $S$ is added to $P[1,3]$, since $D \in P[1,2]$, $E \in P[2,3]$, and $(S \rightarrow D E) \in R$.  Since $S \in P[0,3]$, we conclude that $w$ can be parsed by the grammar.  

\eat{
\gcomment{Revise this part depending on the experiment we choose to present}
Our work aims to demonstrate that it is possible to realize a version of the CYK algorithm using distributed representations, that is, matrices or tensors. More precisely, we show the following two facts.
\begin{itemize}
\item
It is possible to represent both rule set $R$ and 2-dimensional table $P$ as real number matrices $\matrix{R}$ and $\matrix{P}$. In contrast to CYK table $P$, whose size depends on the imput string length, matrix $\matrix{P}$ has fized size.  \gcomment{Mention trade-off with precision ??}
\item
The application of rules in $R$ to table $P$ can be realized by multiplying the two matrices $\matrix{R}$ and $\matrix{P}$.  
\end{itemize}
Figure~\ref{fig:cyk_matrix} also shows a Tetris-like graphical representation of matrix $\matrix{P}_{\mleft}$, which will be discussed in detail later. 
}

\subsection{Distributed Representations with Holographic Reduced Representations}
\label{sec:revisedhrr}

Holographic reduced representations (HRR; \citeauthor{Plate1995}~\citeyear{Plate1995}) are distributed representations well-suited for our aim of encoding the 2-dimensional parsing table $P$ of the CYK algorithm and for implementing the operation of selecting the content of its cells $P[i,j]$. 
\eat{
In fact, HRR can encode symbolic representations and decode them back. 
Moreover, by using holographic reduced representations along with vector shuffling, 
it is possible to encode symbolic structures in distributed representations. Hence, symbols as well as structures in tables $P$ can be encoded and decoded. 
}
In the following, we introduce the operations we use, along with a graphical way to represent their properties. The graphical representation is based on Tetris-like pieces. 

The starting point of a distributed representation is how to encode symbols into vectors: symbol $a$ can be encoded using random vector $\vec{a} \in \R^d$ drawn from a multivariate normal distribution $\vec{a} \sim N(0, \matrix{I}\frac{1}{\sqrt{d}})$. These vectors are used as basis vectors for the Johnson-Lindenstrauss Tranform \cite{Johnson1984} as well as for random indexing \cite{Sahlgren2005}. The major property of these random vectors is the following
\begin{eqnarray*}
\vec{a}^\transpose\vec{b} & \approx & 
\begin{cases}
1 & \text{if } \vec{a} = \vec{b} \\
0 & \text{if }  \vec{a} \neq \vec{b}  \\
\end{cases}
\end{eqnarray*}

Given the above representation of symbols, we can define a basic operation $\sc{\:}$ and its approximate inverse $\invsc{\:}$.  These operations take as input a symbol
and provide a matrix in $\R^{d\times d}$, and are the basis for our encoding and 
decoding.  The first operation is defined as
\begin{eqnarray*}
\sc{a} & = & \cm{A} \Phi \; ,
\end{eqnarray*}
where $\cm{A}$ is the circulant matrix of the vector $\vec{a}$ and $\Phi$ is a permutation matrix. This operation has a nice approximated inverse in 
\begin{eqnarray*}
\invsc{a} & = & \Phi^\transpose \cmt{A} \; . 
\end{eqnarray*}
We then have
\begin{eqnarray*}
\sc{a} \invsc{b} & \approx & \begin{cases}
\matrix{I} & \text{if } \vec{a} = \vec{b} \\
\matrix{0} & \text{if } \vec{a} \neq \vec{b} \\
\end{cases} 
\end{eqnarray*}
since $\Phi$ is a permutation matrix and therefore $\Phi \Phi^\transpose=I$, and since
\begin{eqnarray*}
\cmt{A}\cm{B} & \approx & 
\begin{cases}
\matrix{I} & \text{if } \vec{a} = \vec{b} \\
\matrix{0} & \text{if }  \vec{a} \neq \vec{b}  \\
\end{cases}
\end{eqnarray*}
due to the fact that $\cm{A}$ and $\cm{B}$ are circulant matrices based on random vectors $\vec{a},\vec{b} \sim N(0, \matrix{I}\frac{1}{\sqrt{d}})$.

\eat{
The two operations $\sc{\:}$ and $\invsc{\:}$ are strictly linked to the following notions: (i) the circular convolution and its inverse, called the circular correlation, used to encode and decode flat structures \cite{Plate1995}; and (ii) the shuffled circular convolution \cite{Zanzotto:ICML:2012} used to encode syntactic trees. More precisely, the shuffled circular convolution $\shufcconv$ is defined as 
$$
\vec{a} \shufcconv \vec{b}  = \vec{a} \cconv \Phi \; \vec{b} = \sc{a} \sc{b} \vec{e}_1
$$
where $\cconv$ is the circular convolution  and $\vec{e}_1$ is the first base vector of $\R^d$. Shuffling has been introduced in combination to circular convolution in order to give the possibility of encoding larger structures \cite{Zanzotto:ICML:2012} by eliminating the commutative property of circular convolution. A similar technique has been used for word sequences \cite{Sahlgren2005}.
}

With the $\sc{\;}$ and $\invsc{\;}$ operations at hands, we can now encode and decode strings, that is, finite sequences of symbols. As an example, the string $abc$ can be represented as the matrix product $\sc{a} \sc{b} \sc{c}$.  In fact, we can check that $\sc{a} \sc{b} \sc{c}$ starts with $a$ but not with $b$ or with $c$, since we have $\invsc{a} \sc{a} \sc{b} \sc{c} \approx \sc{b} \sc{c}$, which is different from $\matrix{0}$, while $\invsc{b} \sc{a} \sc{b} \sc{c} \approx \matrix{0}$ and $\invsc{c} \sc{a} \sc{b} \sc{c} \approx \matrix{0}$.  Knowing that $\sc{a} \sc{b} \sc{c}$ starts with $a$, we can also check that the second symbol in $\sc{a} \sc{b} \sc{c}$ is $b$, since $\invsc{b} \invsc{a} \sc{a} \sc{b} \sc{c}$ is different from $\matrix{0}$.  Finally, knowing that $\sc{a} \sc{b} \sc{c}$ starts with $ab$, we can check that the string ends in $c$, since $\invsc{c} \invsc{b} \invsc{a} \sc{a} \sc{b} \sc{c} \approx \matrix{I}$.

Using the above operations, we can also encode sets of strings.  For instance, the string set ${\cal S} = \{abS, DSa\}$ is represented as the sum of matrix products $\sc{a} \sc{b} \sc{S} + \sc{D} \sc{S} \sc{a}$.  We can then test whether $abS \in {\cal S}$ by computing the matrix product $\invsc{S} \invsc{b} \invsc{a}(\sc{a} \sc{b} \sc{S} + \sc{D} \sc{S} \sc{a}) \approx \matrix{I}$, meaning that the answer is positive.  Similarly, $aDS \in {\cal S}$ is false, since $\invsc{S} \invsc{D} \invsc{a}(\sc{a} \sc{b} \sc{S} + \sc{D} \sc{S} \sc{a}) \approx \matrix{0}$.  We can also test whether there is any string in ${\cal S}$ starting with $a$, by computing $\invsc{a}(\sc{a} \sc{b} \sc{S} + \sc{D} \sc{S} \sc{a}) \approx \sc{b} \sc{S}$ and providing a positive answer since the result is different from $\matrix{0}$.

Not only our operations above can be used to encode sets, as described above, they can also be used to encode multi-sets, that is, they can keep a count of the numer of occurrences of a given symbol/string within a collection. For instance, consider a multi-set with two occurrences of symbol $a$.  This can be encoded by means of the sum $\sc{a} + \sc{a}$.  In fact, we can test the number of occurrences of symbol $a$ in the multi-set using the product $\invsc{a}(\sc{a} + \sc{a}) = \matrix{I}+\matrix{I} = 2\matrix{I}$. 

To visualize the encoding and decoding ability of the above operations, we will use a graphical representation based on Tetris. Symbols under the above operations are represented as Tetris pieces: for example, $\sc{a}$ = \symbA{}~, $\invsc{a}$ = \rsymbA{}~, $\sc{b}$ = \symbB{} and $\invsc{b}$ = \rsymbB{}.  In this way strings are sequences of pieces; for example, \symbA\symbB\symbS encodes $abS$ (equivalently, $\sc{a} \sc{b} \sc{S}$).  Then, like in Tetris, elements with complementary shapes are canceled out and removed from a sequence; for example, if \rsymbA is applied to the left of \symbA\symbB\symbS, the result is \symbB\symbS as \rsymbA\symbA disappears.  Sets of strings (sums of matrix products) are represented in boxes, as for instance
\begin{eqnarray*}
B & = & \fbox{\symbA\symbB\symbS \hspace{0.5cm} \symbD\symbS\symbA}
\end{eqnarray*}
which encodes set $\{ abS, DSa \}$ (equivalently, $\sc{a}\sc{b}\sc{S} + \sc{D}\sc{S}\sc{a}$).  In addition to the usual Tetris rules, an element with a certain shape will select from a box only elements with the complementary shape. For instance, if \rsymbA is applied to the left of the above box $B$, the result is the new box
\begin{eqnarray*}
L' & = & \fbox{\symbB\symbS}
\end{eqnarray*}
as \rsymbA selects \symbA\symbB\symbS but not  \symbD\symbS\symbA.

With these operations and with this Tetris metaphor, we can describe our model to encode $P$ tables in matrices and to implement rule applications by means of matrix multiplication, as discussed in the next section.

\begin{figure}
\begin{center}
\setlength{\tabcolsep}{1pt}
\renewcommand{\arraystretch}{1.5}
\begin{tabular}{cccc|ccccc}
\multicolumn{4}{c|}{\emph{numbers}}& \multicolumn{5}{c}{\emph{symbols}} \\
\hline
$\sc{0}$ & $\sc{1}$ & $\sc{2}$ & $\sc{3}$ \; & \; $\sc{a}$ & $\sc{b}$ & $\sc{D}$ & $\sc{E}$ & $\sc{S}$ \\
\indexZero&\indexOne&\indexTwo&\indexThree&\symbA&\symbB&\symbD&\symbE&\symbS\\
\end{tabular}
\caption{Tetris-like graphical representation for the pieces for symbols in our running example.}
\label{tab:tetris_symbols}
\end{center}
\end{figure}

\section{The CYK algorithm on Distributed Representations}

The \termdef{distributed CYK algorithm} (D-CYK) is our version of the CYK algorithm running over distributed representations and using matrix algebra. As the traditional CYK, this algorithm recognizes whether or not  a string $w$ can be generated by a context-free grammar with a set of rules $R$ in Chomsky Normal Form.  Yet, unlike the traditional CYK algorithm, the parsing table $P$ and the rule set $R$ are encoded through matrices in $\R^{d \times d}$, using the distributed representation of Section~\ref{sec:revisedhrr}, and rule application is obtained with matrix algebra.

In this section we describe how the D-CYK algorithm encodes: (i) the table $P$ by means of two matrices $\matrix{P}_{\mleft}$ and $\matrix{P}_{\mright}$; (ii) the unary rules in $R$ by means of a matrix $\rmatrix{u}{A}$ for each nonterminal symbol $A$ in the grammar; and, (iii) binary rules in $R$ by means of matrices $\rmatrix{b}{A}$ for each nonterminal $A$.  We then specify the steps of the D-CYK algorithm and illustrate its execution using the running example of Figure \ref{fig:cyk_matrix}.

\subsection{Encoding the Table $P$ in matrices $\matrix{P}_{\mleft}$ and $\matrix{P}_{\mright}$}

The table $P$ of the CYK algorithm can be seen as a collection of triples $(i,j,X)$. More precisely, the collection of triples contains element $(i,j,X)$ if and only if $X \in P[i,j]$. Given the representation of Section \ref{sec:revisedhrr}, the table $P$ is encoded by means of two matrices $\matrix{P}_{\mleft}$ and $\matrix{P}_{\mright}$ in $\R^{d \times d}$, each containing the collection of triples $(i,j,X)$ in distributed representation.  More precisely, each triple $(i,j,X)$ is encoded as:
\begin{eqnarray*}
\matrix{P}_{\mleft}[i,j,X] & = & \invsc{i} \invsc{j} \invsc{X} \;, \\
\matrix{P}_{\mright}[i,j,X] & = & \sc{X} \sc{i} \sc{j} \;.
\end{eqnarray*}
Then matrix $\matrix{P}_{\mleft}$ is the sum of all elements $\matrix{P}_{\mleft}[i,j,X]$, encoding the collection of all triples from $P$.   Similarly, $\matrix{P}_{\mright}$ is the sum of all elements $\matrix{P}_{\mright}[i,j,X]$. 
To visualize this representation, the matrix $\matrix{P}_{\mleft}$ of our running example is represented in the Tetris-like notation in Figure \ref{fig:cyk_matrix}, where we have used the pieces in Figure \ref{tab:tetris_symbols}.

\subsection{Encoding and Using Unary Rules}
\label{ssec:unary}

The CYK algorithm uses symbols $a_i$ from the input string $w$ and unary rules to fill in cells $P[i-1,i]$, as seen in Algorithm \ref{alg:CYK}.  We simulate this step in our D-CYK using our distributed representation and matrix operations.

D-CYK represents the input $a_i$ using the matrix $\matrix{P}_{\mleft}$.  Hence, before the application of unary rules we have
\begin{eqnarray*}
\matrix{P}_{\mleft} & = & \sum_{i=1}^{n}\invsc{i-1} \invsc{i} \invsc{a_{i}} \; .
\end{eqnarray*}
For processing unary rules we use the first part of the D-CYK algorithm, called D-CYK\_unary and reported in Algorithm \ref{alg:CYK_unary}.  D-CYK\_unary takes $\matrix{P}_{\mleft}$, encoding symbols from the input string, and produces updated matrices $\matrix{P}_{\mleft}$ and $\matrix{P}_{\mright}$ encoding nonterminal symbols resulting from the application of unary rules. 

For the running example of Figure \ref{fig:cyk_matrix}, the initial content of $\matrix{P}_{\mleft}$ is
\begin{equation*}
\matrix{P}_{\mleft} = \fbox{{\centering   
\begin{tabular}{ccc}
\rindexZero{}\rindexOne{}\rsymbA{}
&
\rindexOne{}\rindexTwo{}\rsymbA{}
&
\rindexTwo{}\rindexThree{}\rsymbB{}
\end{tabular}
}}
\end{equation*}
Taking as input the above matrix, D-CYK\_unary produces the two matrices 
\begin{eqnarray*}
\matrix{P}_{\mleft} & = & 
\fbox{{\centering   
\begin{tabular}{ccc}
\rindexZero{}\rindexOne{}\rsymbD
&
\rindexOne{}\rindexTwo{}\rsymbD
&
\rindexTwo{}\rindexThree{}\rsymbE
\end{tabular}
}} \; ,  \\
\matrix{P}_{\mright} & = & 
\fbox{{\centering   
\begin{tabular}{ccc}
\symbD\indexZero{}\indexOne{}
&
\symbD\indexOne{}\indexTwo{}
&
\symbE\indexTwo{}\indexThree{}
\end{tabular}
}}
\end{eqnarray*}

In D-CYK\_unary, we use matrices $\rmatrix{u}{A}$, for each nonterminal $A$ in the left-hand side of some unary rule.  These matrices are conceived to detect the applicability of rules of the form $A \rightarrow a$, where $a$ is some alphabet symbol, to matrix $\matrix{P}_{\mleft}$, and are also used to update matrices $\matrix{P}_{\mleft}$ and $\matrix{P}_{\mright}$.  Matrix $\rmatrix{u}{A}$ is defined as
\begin{eqnarray*}
\rmatrix{u}{A} & = & \sum_{(A \rightarrow a) \in R} \sc{a} \; ,
\end{eqnarray*}
where $R$ is the set of rules of the grammar.  The operation between $\rmatrix{u}{A}$ and $\matrix{P}_{\mleft}$ (line \ref{detection_operation} in Algorithm \ref{alg:CYK_unary}), which detects whether some rule $A \rightarrow a$ is applicable at position $(i-1,i)$ of the input string, is
\begin{eqnarray*}
P_A & = & \sigma(\rmatrix{u}{A} \sc{i} \sc{i-1} \matrix{P}_{\mleft}) \; ,
\end{eqnarray*}
where $\sigma(x)$ is a sigmoid function $\sigma(x) = \frac{1}{1+e^{-(x-0.5)*\beta}}$. 
In fact, $\sc{i}\sc{i-1}\matrix{P}_{\mleft} \approx \invsc{a_i}$ extracts the distributed representation of terminal symbol $a_i$. Then  
\begin{eqnarray}
\label{eq:cases_simple}
\rmatrix{u}{A} \invsc{a_i} & = &
\sum_{(A \rightarrow a)\in R }\sc{a}\invsc{a_i} \approx 
\begin{cases}
\matrix{0} & \text{if } (A \rightarrow a_i) \notin R\\
\matrix{I} & \text{if } (A \rightarrow a_i) \in R\\
\end{cases}
\end{eqnarray}
is reinforced by the subsequent use of the sigmoid function. Hence, if some unary rule with left-hand side symbol $A$ is applicable, the resulting matrix is approximately the identity matrix $\matrix{I}$, else the resulting matrix is approximately the zero matrix $\matrix{0}$. 
Then, the operations in lines \ref{update1} and \ref{update2}: 
\begin{eqnarray*}
\matrix{P}_{\mleft} & = & \matrix{P}_{\mleft} + \invsc{i-1}\invsc{i}\invsc{A}P_A \\
\matrix{P}_{\mright} & = & \matrix{P}_{\mright} + \sc{A}\sc{i-1}\sc{i}P_A 
\end{eqnarray*}
add a non-zero matrix to $\matrix{P}_{\mleft}$ and $\matrix{P}_{\mright}$, respectively, only if rules for $A$ are matched in $\matrix{P}_{\mleft}$ containing the input sentence. 

\begin{algorithm}
\begin{algorithmic}[1]
\State $\matrix{P}_{\mleft} \gets \sum_{i=1}^{n} \invsc{i-1}\invsc{i}\invsc{a_i}$
\For {$i \gets 1$ to $n$}
	\For {$A \in {\mathit \mathit{nonterminals}}$}
	    \State $P_A \gets \sigma(\rmatrix{u}{A}\sc{i} \sc{i-1} \matrix{P}_{\mleft}) $ 
        \label{detection_operation}
    	\State $\matrix{P}_{\mleft} \gets \matrix{P}_{\mleft} + 
        \invsc{i-1}\invsc{i}\invsc{A}P_A $ 
        \label{update1}
    	\State $\matrix{P}_{\mright} \gets \matrix{P}_{\mright} + 
        \sc{A}\sc{i-1}\sc{i}P_A $ 
        \label{update2}
    \EndFor
\EndFor
\end{algorithmic}
\caption{D-CYK\_unary(string $w = a_1 a_2 \cdots a_n$, matrices $\rmatrix{u}{A}$) \Return $\matrix{P}_{\mleft}$ and $\matrix{P}_{\mright}$}
\label{alg:CYK_unary}
\end{algorithm}

We describe the application of D-CYK\_unary using the running example in Figure \ref{fig:cyk_matrix} and the Tetris-like representation. The two unary rules $D \rightarrow a$ and $E \rightarrow b$ are represented as 
$\rmatrix{u}{D}=\sc{a}$ and $\rmatrix{u}{E}=\sc{b}$, that is
$$
\rmatrix{u}{D}= \fbox{ \symbA } \hspace{.5cm}
\rmatrix{u}{E}=\fbox{ \symbB }
$$
in the Tetris-like form. As already seen, given the input sequence $aab$ the matrix $\matrix{P}_{\mleft}$ is initialized as
$$
\matrix{P}_{\mleft} = \fbox{
\begin{tabular}{ccc}
\rindexZero{}\rindexOne{}\rsymbA{}
&
\rindexOne{}\rindexTwo{}\rsymbA{}
&
\rindexTwo{}\rindexThree{}\rsymbB{}
\end{tabular}
}
$$
We focus on the application of rule $D \rightarrow a$ to cell $P[0,1]$ of the parsing table, represented through matrices $\rmatrix{u}{D}$ and $\matrix{P}_{\mleft}$, respectively.  At steps \ref{detection_operation} and \ref{update1} of Algorithm \ref{alg:CYK_unary}, taken together, we have
\begin{align*}
\matrix{P}_{\mleft} &= \matrix{P}_{\mleft} +\rindexZero{}\rindexOne{}\rsymbD\sigma(\fbox{\symbA}\indexOne{}\indexZero{}\matrix{P}_{\mleft}) = \matrix{P}_{\mleft} + \mathit{Update}
\end{align*}
The $\mathit{Update}$ part of the assignment can be expressed as
\begin{align*}
\mathit{Update} &= 
\rindexZero{}\rindexOne{}\rsymbD\sigma(
\fbox{\symbA} \; \indexOne{}\indexZero{}
\fbox{\begin{tabular}{ccc}
  \rindexZero{}\rindexOne{}\rsymbA{} &
  \rindexOne{}\rindexTwo{}\rsymbA{} &
  \rindexTwo{}\rindexThree{}\rsymbB{}
  \end{tabular}}
) \\ &\approx 
\rindexZero{}\rindexOne{}\rsymbD\sigma(
\fbox{\symbA} \; \indexOne{} \rindexOne{}\rsymbA{}) \approx  
\rindexZero{}\rindexOne{}\rsymbD\sigma(\fbox{\symbA{}}\rsymbA{}) \approx 
\rindexZero{}\rindexOne{}\rsymbD 
\end{align*}
This results in the insertion of the distributed representation of element $(0,1,D)$ in $\matrix{P}_{\mleft}$.  A symmetrical operation is carried out to update $\matrix{P}_{\mright}$.  After the application of matrices $\rmatrix{u}{D}$ and $\rmatrix{u}{E}$ at each cell $P[i-1,i]$, the matrices $\matrix{P}_{\mleft}$ and $\matrix{P}_{\mright}$ provide the following values 
\begin{eqnarray}
\matrix{P}_{\mleft} & = & 
\fbox{{
  {\centering  
  \begin{tabular}{cccccc}
  \rindexZero\rindexOne\rsymbD
  &
  \rindexOne\rindexTwo\rsymbD
  &
  \rindexTwo\rindexThree\rsymbE
  &
  \rindexZero{}\rindexOne{}\rsymbA{}
  &
  \rindexOne{}\rindexTwo{}\rsymbA{}
  &
  \rindexTwo{}\rindexThree{}\rsymbB{}
  \end{tabular}}
}} \; ,  
\label{eq:partial_P_left_simple} \\
\matrix{P}_{\mright} & = & 
\fbox{{
  \centering   
  \begin{tabular}{cccccc}
  \symbD\indexZero\indexOne
  &
  \symbD\indexOne\indexTwo
  &
  \symbE\indexTwo\indexThree
  &
  \symbA\indexZero\indexOne
  &
  \symbA\indexOne\indexTwo
  &
  \symbB\indexTwo\indexThree
  \end{tabular}
}}  \; .
\label{eq:partial_P_right_simple}
\end{eqnarray}

\subsection{Encoding and Using Binary Rules}
\label{ssec:binary}

To complete the specification of algorithm D-CYK, 
we describe here how to encode binary rules in such a way
that these rules can be fire over the distributed representation of table $P$ 
through operations in our matrix algebra. 
We introduce the second part of the algorithm, called D-CYK\_binary 
and specified in Algorithm~\ref{alg:CYK_binary}, and we clarify why 
both $\matrix{P}_{\mleft}$ and $\matrix{P}_{\mright}$, introduced
in section~\ref{ssec:binary}, are needed.

Binary rules in $R$ with nonterminal symbol $A$ in the left-hand side
are all encoded in a matrix $\rmatrix{b}{A}$ in $\R^{d \times d}$.
$\rmatrix{b}{A}$ is conceived for defining matrix operations that 
detect if rules of the form $A \rightarrow BC$, for some $B$ and $C$, 
fire in position $(i,j)$, given $\matrix{P}_{\mleft}$ and $\matrix{P}_{\mright}$. 
This operations result in a nearly identity matrix $\matrix{I}$ 
for a specific position $(i,j)$ if at least one specific rule coded 
in $\rmatrix{b}{A}$ fires in position $(i,j)$ over positions $(i,k)$ 
and $(k,j)$, for any value of $k$. This will enable the insertion of new symbols 
in $\matrix{P}_{\mleft}$ and $\matrix{P}_{\mright}$.

\begin{algorithm}[H]
\begin{algorithmic}[1]
\For {$j \gets 2$ to $n$}
	\For {$i \gets j-2 $ to $0$} 
    		\For {$A \in \mathit{nonterminals}$}
             \State $P_A \gets \sigma(\sc{j}\invsc{i} \tensor{P}_{\mleft} \rmatrix{b}{A} \tensor{P}_{\mright}) \otimes \matrix{I}$ \label{cool_idea}
             \State $\tensor{P}_{\mleft} \gets \tensor{P}_{\mleft} + \invsc{i}\invsc{j}\invsc{A}P_A$
             \State $\tensor{P}_{\mright} \gets \tensor{P}_{\mright} + \sc{A}\sc{i}\sc{j}P_A$
    		\EndFor
    	\EndFor
\EndFor
\end{algorithmic}
\caption{D-CYK\_binary($\matrix{P}$, rules $R_A$ for each $A$) \Return $\matrix{P}$}
\label{alg:CYK_binary}
\end{algorithm}

To define $\rmatrix{b}{A}$ we encode the right-hand side of 
each binary rule $A \rightarrow BC$ in $R$ as 
\begin{eqnarray*}
\matrix{r}_{A \rightarrow B C} & = & \invsc{B}\invsc{C}
\end{eqnarray*}
All the right-hand sides of binary rules with symbol $A$ in the left-hand side
are then collected in matrix $\rmatrix{b}{A}$
\begin{eqnarray*}
\rmatrix{b}{A} & = & \sum_{A \rightarrow B C} \invsc{B}\invsc{C}. 
\end{eqnarray*}
Algorithm \ref{alg:CYK_binary} uses these rules to determine whether a symbol $A$ can fire in a position $(i,j)$ for any $k$. The key part is line \ref{cool_idea} of Algorithm \ref{alg:CYK_binary}
$$
P_A = \sigma(\sc{j}\invsc{i} \tensor{P}_{\mleft} \rmatrix{b}{A} \tensor{P}_{\mright}) \otimes \matrix{I} \; .
$$
$\rmatrix{b}{A}$ selects elements in $\tensor{P}_{\mleft}$ and $\tensor{P}_{\mright}$ 
according to rules for $A$.  Matrices $\tensor{P}_{\mleft}$ 
and $\tensor{P}_{\mright}$ have been designed in such a way that, 
after the nonterminal symbols in the selected elements have been annihilated,  
the associated spans $(i,k)$ and $(k,j)$ merge into span $(i,j)$.  
Finally, the terms $\sc{j}\invsc{i}$ are meant to check 
whether the span $(i,j)$ has survived. If this is the case, the resulting matrix will be 
very close to the identity matrix $\matrix{I}$, otherwise it will be 
very close to the null matrix $\matrix{0}$.  To reinforce this matrix, 
similarly to Algorithm \ref{alg:CYK_unary}, we use the sigmoid function 
$\sigma(x)$. 
Finally, we apply an element-wise multiplication with $\matrix{I}$, 
which is helpful to remove noise.

To visualize the behavior of the algorithm D-CYK\_binary and the effect of using $\rmatrix{b}{A}$, we use again the running example of Figure \ref{fig:cyk_matrix}. 
The only nonterminal with binary rules is $S$, and matrix $\rmatrix{b}{S}$ is:
$$\rmatrix{b}{S} = \fbox{\symbD\rsymbE \hspace{0.1cm} \symbD\rsymbS}$$
Let's focus on the position $(1,3)$. The operation is the following:
$$
P_A =  \sigma(\fbox{\rindexThree\indexOne } \; \tensor{P}_{\mleft} \; \fbox{\symbD\rsymbE \hspace{0.1cm} \symbD\rsymbS} \; \tensor{P}_{\mright})\otimes \matrix{I} 
$$
where $\tensor{P}_{\mleft}$ and $\tensor{P}_{\mright}$ are as 
in equations~\ref{eq:partial_P_left_simple} 
and~\ref{eq:partial_P_right_simple}. We can then write
$$ 
\fbox{\rindexThree\indexOne } \; \tensor{P}_{\mleft} \; 
\fbox{\symbD\rsymbE \hspace{0.1cm} \symbD\rsymbS} \; 
\tensor{P}_{\mright} =  
$$
$$ 
\fbox{\rindexThree\indexOne } \; 
\fbox{\rindexZero\rindexOne\rsymbD
  \hspace{0.1cm}
  \rindexOne\rindexTwo\rsymbD
  \hspace{0.1cm}
  \rindexTwo\rindexThree\rsymbE \hspace{0.1cm} $\ldots$}  \; 
\fbox{\symbD\rsymbE \hspace{0.1cm} \symbD\rsymbS} \; 
\fbox{\symbD\indexZero\indexOne
  \hspace{0.1cm}
  \symbD\indexOne\indexTwo
  \hspace{0.1cm}
  \symbE\indexTwo\indexThree
  \hspace{0.1cm} $\ldots$} \; \approx 
$$
$$ 
\fbox{\rindexThree\indexOne } \; 
\fbox{\rindexZero\rindexOne\rsymbD \symbD\rsymbE
  \hspace{0.1cm}
  \rindexOne\rindexTwo\rsymbD \symbD\rsymbE
  \hspace{0.1cm}
  \rindexZero\rindexOne\rsymbD \symbD\rsymbS
  \hspace{0.1cm}
  \rindexOne\rindexTwo\rsymbD \symbD\rsymbS}  \; 
\fbox{\symbD\indexZero\indexOne
  \hspace{0.1cm}
  \symbD\indexOne\indexTwo
  \hspace{0.1cm}
  \symbE\indexTwo\indexThree
  \hspace{0.1cm} $\ldots$} \; \approx 
$$
$$ 
\fbox{\rindexThree\indexOne } \; 
\fbox{\rindexZero\rindexOne\rsymbE
  \hspace{0.1cm}
  \rindexOne\rindexTwo\rsymbE
  \hspace{0.1cm}
  \rindexZero\rindexOne\rsymbS
  \hspace{0.1cm}
  \rindexOne\rindexTwo\rsymbS}  \; 
\fbox{\symbD\indexZero\indexOne
  \hspace{0.1cm}
  \symbD\indexOne\indexTwo
  \hspace{0.1cm}
  \symbE\indexTwo\indexThree
  \hspace{0.1cm} $\ldots$} \; \approx 
$$
$$ 
\fbox{\rindexThree\indexOne } \; 
\fbox{\rindexZero\rindexOne\rsymbE \symbE\indexTwo\indexThree
  \hspace{0.1cm}
  \rindexOne\rindexTwo\rsymbE \symbE\indexTwo\indexThree} \; \approx \; 
\fbox{\rindexThree\indexOne } \; 
\fbox{\rindexZero\rindexOne\indexTwo\indexThree
  \hspace{0.1cm}
  \rindexOne\rindexTwo\indexTwo\indexThree} \; \approx
$$
$$ 
\fbox{\rindexThree\indexOne } \; 
\fbox{\rindexZero\rindexOne\indexTwo\indexThree
  \hspace{0.1cm}
  \rindexOne\indexThree} \; \approx \;
\fbox{\rindexThree\indexOne \rindexOne\indexThree} \; \approx \;
\fbox{\rindexThree\indexThree} \; \approx \matrix{I}
$$
Then, this operation detects that $S$ is active in the position $(1,3)$.

With this second part of the algorithm, CYK has been reduced to D-CYK, which works with distributed representations and matrix operations compliant with neural networks.

\begin{figure*}
\centering
\begin{subfigure}{.3\textwidth}
  \centering
  \includegraphics[width=5cm]{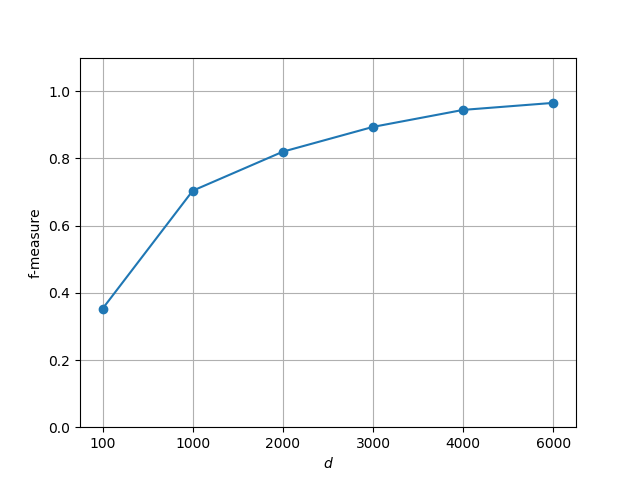}
  \caption{f-measures vs. dimensions $d$ of $\R^d$  on sentences with length $\leq 7$ and the grammar $G_0$}
  \label{fig:sfig1}
\end{subfigure}%
\hspace{0.1cm}
\begin{subfigure}{.3\textwidth}
  \centering
	\includegraphics[width=5cm]{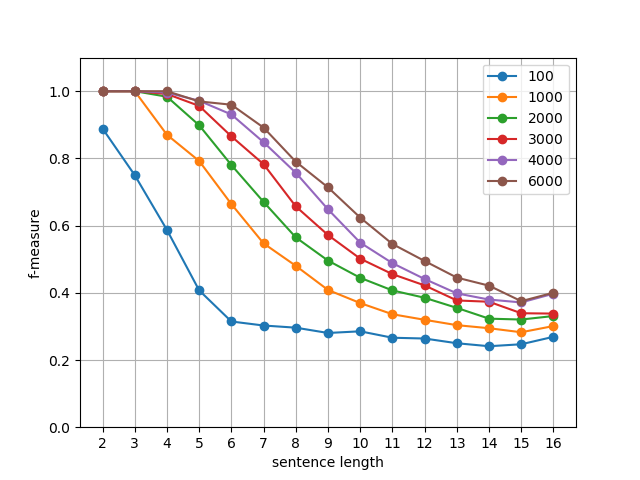}
  \caption{f-measures vs. lengths of sentences with different dimensions $d$ of $\R^d$ on the grammar $G_0$}
  \label{fig:sfig2}
\end{subfigure}
\hspace{0.1cm}
\begin{subfigure}{.3\textwidth}
  \centering
	\includegraphics[width=5cm]{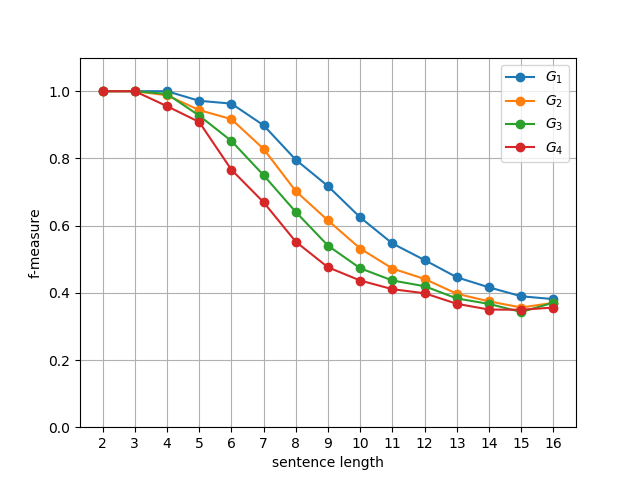}
  \caption{f-measures vs. lengths of sentences with different grammars ($G_1$, $G_2$, $G_3$ and $G_4$)}
  \label{fig:sfig3}
\end{subfigure}%
\caption{Cell f-measures for different configurations of the Distributed CYK vs. the CYK}
\label{fig:results}
\end{figure*}

\section{Experiments}

The aim of these experiments is to show that the distributed CYK algorithm behaves as the original CYK algorithm. For this purpose, we do not need huge datasets but small well-defined set of sentences derived from fixed grammars as defined in the following sections.

\subsection{Experimental Set-up}

We experimented with five different grammars with increasing sets of rules and a set of 2,000 sentences. The grammars are: $G_0$ is the basic grammar with $8$ rules with $3$ unary rules; $G_1$ with $25$ rules expands only unary rules over $G_0$; $G_2$ with $28$ rules expands only binary rules over $G_1$;  $G_3$ with $34$ rules expands only binary rules over $G_1$; and, finally, $G_4$ with $41$ rules expands only binary rules over $G_1$.
The set of sentences has been produced by randomly generated 2,000 sentences of different lengths by using the grammar $G_0$. These sentences can be recognized by all the 5 grammars as all the other grammars are obtained by adding rules to $G_0$.



As we want to understand whether D-CYK is able to reproduce the computation of the original CYK,  we used \termdef{cell f1-measure} ($f1$), which evaluates whether the distributed P is similar to the original P. To evaluate $f1$, we decoded the distributed versions $\matrix{P}_{\mleft}$ of $P$ by using a simple decoding algorithm $Dec$ (Algorithm \ref{alg:decoder}).
\begin{algorithm}
\begin{algorithmic}[1]
\For {$i \gets 0$ to $n$}
	\For {$j \gets i+1 $ to $n+1$}
    		\For {each $A$ in $R$}
				\If {$\sigma(\sc{A}\sc{j}\sc{i}\matrix{P}_{\mleft})_{0,0}$ > 0.99} \label{l:binary}
					\State add $A$ to $P[i,j]$  
				\EndIf
    		\EndFor
    \EndFor
\EndFor
\end{algorithmic}
\caption{Dec($\matrix{P}_{\mleft}$) \Return table $P$}
\label{alg:decoder}
\end{algorithm}
We compared $Dec(\matrix{P}_{\mleft})$ and the matrix $P$ obtained by applying the traditional CYK algorithm with the corresponding grammar on test sentences. By comparing $Dec(\matrix{P}_{\mleft})$ and the corresponding $P$, we evaluated the \termdef{cell precision} and the \termdef{cell recall} by considering $P$ the oracle and $Dec(\matrix{P}_{\mleft})$ the system. $f1$ is then computed according to the traditional equation of mixing precision and recall.    


We experimented with six different dimensions of matrices $\matrix{P}$: 100, 1000, 2000, 3000, 4000, 5000  and 6000.

\subsection{Results}

Results are really encouraging showing that, as the dimensions of the matrices increase, D-CYK can approximate with its operations what is done by the traditional CYK. The f1-measure is in fact increasing with the dimension of the matrix. This is mainly due to an improvement of the \termdef{cell symbol precision} as the \termdef{cell symbol recall} is substantially stable. Hence, as the dimension increases D-CYK gets more precise in replicating the original matrix.

The size of the grammar is instead a major problem. In fact, the precision of the algorithm is affected by the number of rules whereas the recall is substantially similar across the three different grammars. 

These results confirm that it is possible to transfer a traditional algorithm on a version, which is defined on distributed representations.

%
%

\section{Conclusions and Future Work}

In these days, the predominance of symbolic, grammar-based syntactic parsers for natural language has been successfully challenged by neural networks, which are based on distributed representations. Years of results and understanding can be lost.  We proposed D-CYK that is a distributed version of the CYK, a classical parsing algorithm. Experiments show that D-CYK can do the same task of the original CYK in this new setting. 

Neural networks are a tremendous opportunity to develop novel solutions for known tasks.  Our solution opens an avenue to an innovative set of possibilities: revitalizing symbolic methods in neural networks. In fact, our algorithm is the first step towards the definition of a ``complete distributed CYK algorithm'' that builds trees in distributed representations during the computation. Moreover, it can foster the definition of recurrent layers of CYK-informed neural networks.

\starttwocolumn
\bibliographystyle{compling}
\bibliography{MyCollection}

\end{document}